# Multi-Label Zero-Shot Learning via Concept Embedding

Ubai Sandouk and Ke Chen

**Abstract**—Zero Shot Learning (ZSL) enables a learning model to classify instances of an unseen class during training. While most research in ZSL focuses on single-label classification, few studies have been done in multi-label ZSL, where an instance is associated with a set of labels simultaneously, due to the difficulty in modeling complex semantics conveyed by a set of labels. In this paper, we propose a novel approach to multi-label ZSL via concept embedding learned from collections of public users' annotations of multimedia. Thanks to concept embedding, multi-label ZSL can be done by efficiently mapping an instance input features onto the concept embedding space in a similar manner used in single-label ZSL. Moreover, our semantic learning model is capable of embedding an out-of-vocabulary label by inferring its meaning from its co-occurring labels. Thus, our approach allows both seen and unseen labels during the concept embedding learning to be used in the aforementioned instance mapping, which makes multi-label ZSL more flexible and suitable for real applications. Experimental results of multi-label ZSL on images and music tracks suggest that our approach outperforms a state-of-the-art multi-label ZSL model and can deal with a scenario involving out-of-vocabulary labels without re-training the semantics learning model.

## 1 INTRODUCTION

Zero-Shot Learning (ZSL) refers to a task that establishes a learning model which can classify instances of an unseen class during learning, named ZSL-class, with only training examples of seen classes, dubbed T-classes hereinafter. ZSL increases the capacity of a classifier in dealing with a situation where ZSL-class training examples are unavailable [1]. The main idea behind ZSL [2] is associating T-classes with ZSL-classes semantically via the use of additional knowledge on meaning of different class labels (normally in a specific domain) to form a uniform semantic representation for ZSL- and T-classes. Then, a mapping function from input data onto the semantic representation of T-classes is established via learning. In test, this mapping function is applied to an unknown instance to predict the semantic representation of its ground-truth label in ZSL- or T-classes. Finally, a ZSL-class label derived from its predicted semantic representation is assigned to this testing instance. Based on the aforementioned idea, several ZSL approaches have been proposed for single-label classification [2]–[5], where any instance is merely associated with a single class label. Single-label ZSL approaches have been successfully applied to real world problems, e.g., fMRI brain scan interpretation [6], textual query intention categorization [7], and object recognition [3].

In reality, an instance may be associated with a set of class labels simultaneously, which results in multi-label classification [8]. For example, an image often contains a number of different objects as well as a background; and hence, needs to be described with several labels together. As pointed out in [8], multi-label classification is a more difficult task than single-label classification. It is of great importance to extend ZSL to multi-label classification as is required by multimedia information processing. However, multi-label ZSL has to address some issues that do not exist in single-label ZSL. To a large extent, multi-label ZSL remains an open problem [9], mainly due to the complex underlying corresponding relationship between an instance and a set of labels used to describe it.

In general, there are two challenging problems in multi-label ZSL; i.e., a) how to create a semantic representation that properly encodes the entire complex semantics conveyed in a set of labels; and b) how to map an instance to this semantic representation involving a set of multiple labels. Apparently, a solution to the latter problem entirely depends on the outcome of the former. Therefore, an effective solution to modeling the complex semantics is absolutely crucial for the success of multi-label ZSL. However, modeling semantics for multi-label ZSL is quite distinct from that for single-label ZSL.

In single-label ZSL, each label can be uniquely represented in a semantics space; in other words, the meaning of a label and the relatedness between two different labels are all fixed. In this paper, we refer to such semantics as *global semantics*. To obtain a global semantic representation, there are two approaches in general: manually converting a label into a list of pre-defined attributes that can characterize all possible labels in a specific domain [5], and automatically learning a continuous semantic embedding space from linguistic resources, e.g., semantic embedding learning from Wikipedia leads to the well-known word2vec space [10], [2].

In contrast, multi-label ZSL involves sets of labels that convey complex semantics, e.g., polysemantic aspect of a label and collective semantics reflecting different concepts. For example, two image instances are annotated with sets of labels: {"apple", "mobile", "phone", "5s"} and {"apple", "knife", "kitchen"}, respectively. Obviously, "apple" in the former means the company that produces a brand mobile phone while the latter refers to a kind of fruit. Apparently, a specific meaning of "apple" remains uncertain unless other co-occurring labels in the set are seen. Furthermore, each label reflects a concept and all the co-occurring labels in a set collectively convey the semantics, e.g., {"apple", "mobile", "phone", "5s"} together indicate "iPhone 5s", while {"apple", "knife", "kitchen"} collectively express an indoor scenery. Instead of a global semantic representation, a proper semantic representation is required for multi-label ZSL via modeling the complex semantics that is referred to as *contextualized semantics* in this paper. Nevertheless, most of existing approaches to modeling semantics underlying a set of labels do not meet the requirement of a contextualized semantic representation. On one hand, statistical semantics modeling techniques, such as latent Dirichlet allocation [11] and conditional restricted Boltzmann machines [12], only yield compact statistical summaries of groups of labels which means that such techniques are confined to capturing the most probable patterns of label co-occurrence ignoring label inter-relatedness. On the other hand, distributed linguistic models, e.g., [10], [13], work under the condition that there is syntactic relatedness between words but a set of labels does not comply with this condition.

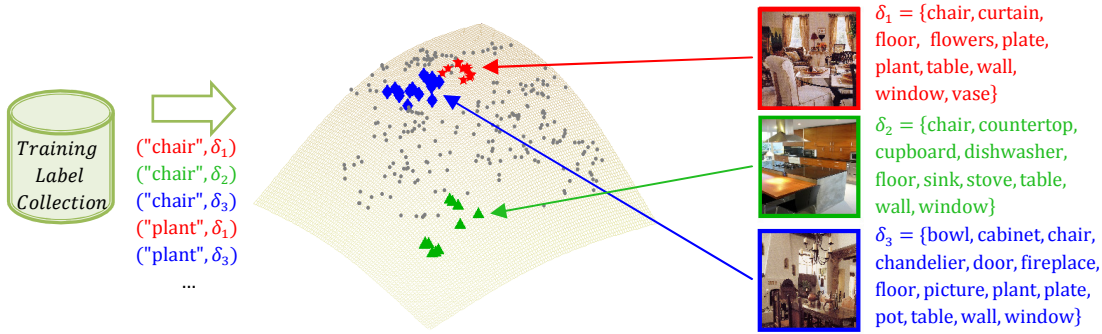

Fig.1. The proposed *Concept Embedding* approach for multi-label ZSL. The notation ("x", δ) stands for label x in context of δ. Annotated image instances are from HSUN [14]. A set of ground-truth labels used to describe each image is listed along with the image.

In ZSL, there is another issue that has not been addressed adequately; i.e., some labels used to annotate instances are beyond a vocabulary of pre-defined labels in modeling semantics [4], [15]. Hereinafter, we dub such labels *out-of-vocabulary* (OOV) labels. The presence of OOV labels poses a challenge in establishing a mapping from an instance to its corresponding semantic representation. To the best of our knowledge, this issue was only addressed inadequately by either adding OOV labels to the pre-defined vocabulary or simply abandoning such training examples during learning the mapping. The former has to model semantics again from scratch, which is time-consuming and might require more data, while the latter inevitably incurs information loss.

To tackle problems arising from multi-label ZSL, a few attempts have been made. The work in [16] uses the compositionality properties of word2vec space [17] in order to achieve collective representation of labels. However, annotating an instance requires exhaustive search within all label combinations, which results in a prohibitive deployment complexity. To overcome this weakness, the work in [9] proposes a multi-instance semantic embedding for multi-label ZSL in the image domain where each individual patch containing a single object is mapped onto a semantic representation similar to single-label ZSL.

However, this approach can only be applied to images by assuming that patches containing individual objects can always be identified. Unlike the above approaches, the work in [18] suggests the use of co-occurrence statistics among training and ZSL labels. Although this model uses semantics obtained from labels, it ignores the correlation between labels since it independently predicts labels one by one. In general, existing multi-label ZSL approaches are either limited to a specific domain [9] or subject to technical limitations [16], [18].

In this paper, we propose a novel approach to multi-label ZSL based on our latest work [19]. We fight off the multi-label ZSL challenges via two stages. Fig. 1 illustrates the basic idea underlying our approach. We assume that a label along with its co-occurring labels in a label set describing an instance formulate a specific concept. In the first stage, we learn concept embedding (CE) via a semantic training dataset that contains sets of coherent labels used to describe instances in a domain. Thus, a label has polysemantic representations as it is co-occurring with different labels (in different sets of labels) and the Euclidean distance between embedded concepts in the CE space reflects their semantic similarity. In Fig. 1, a concept denoted by ("x", δ) is seen as • in the CE space. For example, the label "chair" in context $\delta_1$ and in context $\delta_2$ defines two different concepts which we highlight separately using ★ and ▲. Furthermore, a set of co-occurring labels frame a number of similar concepts and hence their embeddings are co-located or close together, e.g., all the concepts defined by 10 labels describing the image *modern dining room*, i.e., $\delta_1$, are co-located as 10 ★s. In the second stage, we learn mapping of instances onto the CE space via the set of labels used to describe them. By using such a mapping, all the labels related to a test instance can be identified easily, e.g., three real image instances in Fig. 1.

Overall, the main contributions of this study are in two aspects: a) we present a generic multi-label ZSL framework that can deal with a number of challenging problems including concept embedding regardless of application domains, semantic modeling of OOV labels without need of re-training the semantic learning model and a novel manner for efficiently establishing a mapping from an instance to its CE representation; and b) We demonstrate that the CE space learned from co-occurring labels is effective in multi-label ZSL as our approach outperforms a state-of-the-art multi-label ZSL in both image and music domains with different experimental settings.

The remainder of this paper is organized as follows: Sect. 2 briefly lists related works. Sect. 3 presents our CE based multi-label ZSL framework. Sect. 4 describes the experimental design and settings, and Sect. 5 reports experiential results. Sect. 6 discusses issues arising from this study, and the last section draws conclusions.

## 2 RELATED WORKS

In this section, we briefly outline connections and main differences to existing multi-label ZSL approaches.

The successful use of linguistic word embedding spaces, e.g., word2vec [10] and GloVe [13], in single-label ZSL [2], [4] encouraged extending previous works into the multi-label case. As a result, the challenge of learning semantics is overlooked. However, mapping instances onto such spaces is challenging. In [16], all known labels are represented as vectors and the compositionality of word2vec space [17] is directly used. The set of labels associated with a training instance are collected to obtain an instance level representation based on the assumption that these labels have similar compositionality properties as English words in the semantics space. As a result, a mapping is learned from an instance to a "compressed" representation of its associated labels by summing up the semantic representations of these labels [16]. Due to a lack of proper semantic representations, [16] requires an exhaustive search over all combinations of labels, which is computationally prohibitive when there are a large number of labels. In fact, [16] used only test datasets of up to eight labels in their experiments.

The work in [9] adopted GloVe [13] to label individual image patches where all known labels are represented as vectors. Thus, semantically meaningful patches in an image are identified by geodesic object proposals [20] and then individually mapped to vectors of their ground-truth labels in a semantics space. This model assumes that meaningful image patches can always be obtained where each patch contains a single object. However, there are labels that describe entire images instead of single objects and a patch may be annotated with more than one label. Furthermore, small objects might be overlooked or misclassified when there are many objects in an image [21]. This approach [9] is not extensible to other domains, e.g., it is extremely difficulty to segment a music track into semantically coherent pieces where each piece can be labeled with a single label.

In general, approaches in [9], [16] rely on linguistic semantics that only concerns words but neglect exploration of label correlation semantics. Overcoming these weaknesses and limitations demand learning semantics that is native to multi-label ZSL. As a result, the *Co-Occurrence Statistics for Zero-Shot Classification* (COSTA) model [18] was proposed by exploring contextualized label co-occurrence. COSTA employs a linear model that predicts the suitability of a ZSL label based on the predicted training labels. As a result, the challenge of learning semantics is addressed by observing co-occurrence of training and ZSL labels in a semantics learning dataset. Subsequently, learning the mapping from instances to the label semantics representation is boiled down to multi-label classification over training labels [18]. While COSTA can directly benefit from state-of-the-art multi-label classification techniques, its ZSL predictions are simply a direct extension of predicted training labels resulting from a multi-label classifier. Nevertheless, COSTA learns native semantics from label collections although it still neglects the correlation between labels. In contrast to other models [9], [16], COSTA is closest to our proposed approach.

In summary, the existing multi-label ZSL approaches are subject to various technical limitations and almost all previous works are in the image domain, e.g., [9], [16], [18]. In this paper, we propose a novel yet generic approach to overcome these limitations and to be applied in different application domains. In particular, it is the first time that an approach addresses the OOV issue in context of multi-label ZSL.

## 3 CONCEPT EMBEDDING BASED MULTI-LABEL ZSL

In this section, we present our *concept embedding based multi-label ZSL* (CE-ML-ZSL) framework. We first describe our problem statement and main idea. Then, we present our technical solutions in detail.

### 3.1 Overview

The multi-label ZSL is to learn a mapping $f: \mathcal{R}^{d^{(I)}} \to [0,1]^{|\Gamma|}$, where the input $x \in \mathcal{R}^{d^{(I)}}$ is the instance characterized by $d^{(I)}$ features, and the output $l \in [0,1]^{|\Gamma|}$ is a list of $|\Gamma|$ ranked label-relatedness scores for $x$. Here, $\Gamma = \Gamma^{(T)} \cup \Gamma^{(ZSL)}$ is a vocabulary containing both T-class labels in $\Gamma^{(T)}$ and ZSL-class labels in $\Gamma^{(ZSL)}$, but no training examples of ZSL-class labels are available when learning the mapping $f$.

As pointed out previously, it is essential to address two challenging issues in multi-label ZSL: finding out a proper semantic representation concerning the complex semantics underlying a set of labels drawn from a predefined label vocabulary $\Gamma$; and b) establishing a mapping from an instance to this semantic representation regarding a set of labels used to describe this instance. In our approach, we tackle these two issues by formulating them as two subsequent learning problems.

In order to find a proper semantic representation to model the complex semantics conveyed by a set of labels, we formulate it as a *concept embedding* (CE) problem [19]: $CE: \Gamma \times \Delta \to \mathcal{R}^{d^{(S)}}$ where $\Delta$ is a domain-dependent collection containing all the sets of labels used to annotate instances. For a set of co-occurring labels, $\delta = \{\tau_i\}_{i=1}^{|\delta|}$ where $\tau_i \in \Gamma$ and $\delta \in \Delta$, it is assumed that $\tau_i$ along with its co-occurring labels in $\delta$ (all the labels in $\delta$ collectively are named *local context* for any label in $\delta$ hereinafter) defines a specific concept. Thus, a label in different local contexts formulates different concepts. As a result, a label has multiple CE representations in different local contexts. Moreover, Euclidean distance between concepts in CE space reflects their semantic similarity (for intuition, see the CE examples shown in Fig. 1). The CE representations capture the contextualized semantics and polysemantic aspects of a label. Hence, the collective use of CE representations derived from a set of coherent labels would accurately model the complex semantics underlying the set of labels as required by multi-label ZSL. To carry out the CE, we proposed a Siamese neural architecture and trained it with a semantics learning dataset of a predefined vocabulary $\Gamma^{(S)}$ [19], to be described in Sect. 3.2. As illustrated in Fig. 2, after the CE learning, we obtain a mapping $CE$ that yields continuous semantic representations for concepts defined by labels along with their local contexts in $\Delta$ where all $\sum_{\tau \in \Gamma^{(S)}} N_\tau^{(S)}$ known concepts resulting from the semantic learning dataset are highlighted in the CE space $S$ of $d^{(S)}$ dimensions where $N_\tau^{(S)}$ is the number of label sets containing label $\tau$.

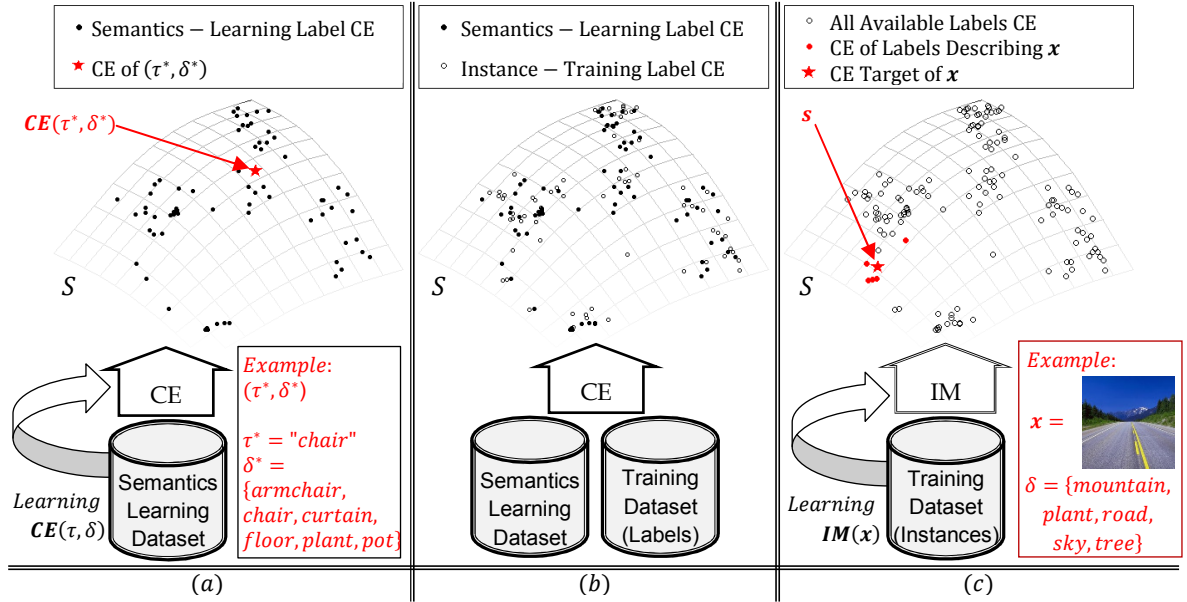

Fig. 2. The *CE-ML-ZSL* framework. (a) Concept embedding learning with a semantics learning dataset. (b) Concept embedding (CE) with the learned CE model. (c) Instance mapping (IM) learning with a multi-label instance training dataset.

To establish a mapping from an instance to the CE semantics representation regarding a set of labels used to describe this instance, we employ an instant training dataset to learn such a mapping based on the output of the CE model. However, we encounter two challenging problems; i.e., the OOV labels and the variable number of labels used in describing different instances. Due to two subsequent learning stages, the vocabulary $\Gamma^{(T)}$ in the instance mapping learning may contain labels beyond the vocabulary $\Gamma^{(S)}$ in reality, which leads to the OOV problem. Due to a variable number of labels used to describe different instances, the existing methods [9], [16], [18] have computational limitations in learning a mapping $\mathbb{f}$ to yield a list of $|\Gamma|$ ranked label-relatedness scores for an instance especially when there is a large number of labels in $\Gamma$, as reviewed in Sect. 2.

To address the OOV issue, we use a method proposed in our previous work [19] based on the nature of our CE space. As a result, an OOV-label related CE representation can be inferred from those of its co-occurring labels used to describe an instance, to be described in Sect. 3.3. Once the OOV issue is addressed, concepts defined by all sets of labels describing instances (in a training dataset) would be properly embedded in the CE space. The $\sum_{\tau \in \Gamma^{(T)}} N_\tau^{(T)}$ added known concepts arising from sets of labels in the instance training dataset are highlighted in Fig. 2(b) for illustration, where $N_\tau^{(T)}$ is the number of label sets involving label $\tau$.

Instead of learning a mapping $\mathbb{f}$ directly, we formulate an alternative learning problem: $IM: \mathcal{R}^{d^{(I)}} \to S$ by means of the CE nature; i.e., similar concepts defined by a set of co-occurring labels are co-located or close to one another in CE space. Instead of using all CE representations derived from a set of labels used to describe an instance, we set the target in this learning task to a "compressed" CE representation, $s \in S$, which collectively summarizes all the concepts formulated by the set of labels. Thus, the $IM$ learning, to be presented in Sect. 3.3, is not affected by the varying number of labels in a set used to describe an instance. Fig. 2(c) illustrates the $IM$ learning process where for an instance, $(x, \delta)$, the CE representations of labels in $\delta$ and the target derived from labels in $\delta$ are highlighted.

In application, the target CE representation of a test instance $\hat{x}$ is predicted: $\hat{s} = IM(\hat{x})$. However, this result does not reach the ultimate goal of multi-label ZSL, a list of $|\Gamma|$ ranked label-relatedness scores for $\hat{x}$. Thanks to the nature of our CE space, generating the list of ranked scores for all the labels in $\Gamma$ can be converted into semantic priming [22], a well-known task in information retrieval. By using semantic priming, the ultimate goal is attained by measuring distances between $\hat{s}$ and all known concepts to generate $|\Gamma|$ ranked label-relatedness scores with a simple algorithm, to be presented in Sect. 3.4. Hence, the ranked scores of all the labels in $\Gamma$ are achieved efficiently. Fig. 3 illustrates the application process of our CE-ML-ZSL approach via an example. As illustrated in Fig. 3(a), concepts of increasing distance away from $\hat{s}$ have less relatedness to $\hat{x}$. The top scores in $\hat{l}$ achieved via semantic priming are listed in Fig. 3(b).

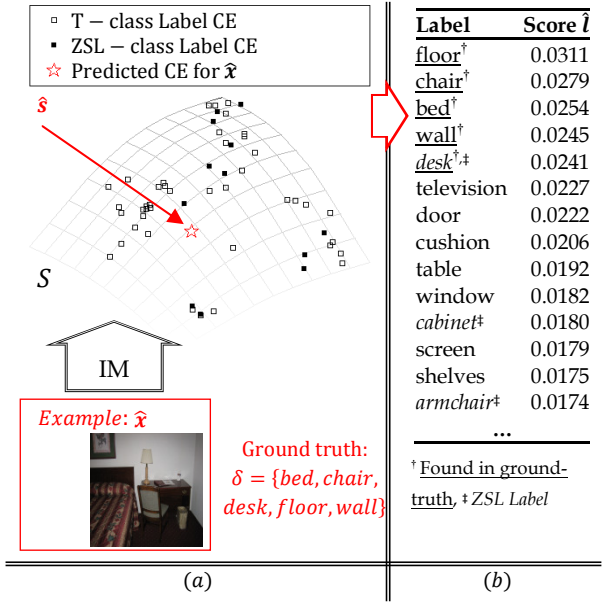

Fig. 3. A *CE-ML-ZSL* application exemplification. (a) Prediction of target CE $\hat{s}$ for a test instance $\hat{x}$ via the *IM* model (the ground-truth is shown for reference) and subsequent semantic priming. (b) The resultant scores of top related labels assigned to $\hat{x}$.

### 3.2 Concept Embedding Learning

To be self-contained, we briefly describe our approach to learning $CE: \Gamma \times \Delta \rightarrow \mathcal{R}^{d^{(S)}}$ developed in our very recent work and more details can be found in [19].

#### 3.2.1 Label, Context and Document Representation

Our CE learning approach [19] is based on raw label, context and document representations.

A label $\tau \in \Gamma^{(S)}$ is described by analyzing its global pattern of usage in a semantics learning dataset via aggregation [23]. As a result, the $tfidf$ weights of each label's use are first extracted to highlight rare but informative labels. Then, dot product on pairs of labels' uses are applied to uncover pair-wise shared patterns of use. Finally, each label is described by its shared pattern of use against all other labels in the training set. The resulting feature vector $t(\tau)$ is of dimensionality $|\Gamma^{(S)}|$ and summarizes the global use of each label.

The local context of a label $\tau$, formed by a document $\delta$, a set of co-occurring labels, is captured via Latent Dirichlet Allocation (LDA) [11] that characterizes the local context with a histogram over a set of latent topics $\Phi$ as $lc(\delta)$, leading to a representation of $|\Phi|$ features.

To facilitate the proposed learning cost function, the Bag-of-Words, $BoW(\delta)$ is also employed to represent a document $\delta$ via a sparse feature vector of $|\Gamma^{(S)}|$ entries.

#### 3.2.2 Siamese Neural Architecture

For CE learning, we proposed a Siamese neural architecture where a deep neural network was used as a component sub-network. As depicted in Fig. 4, a sub-network consists of $H$ consecutive layers of nonlinear units and is fed with the input: $x^{(S)}(\tau, \delta) = \{t(\tau), lc(\delta)\}^1$ formed by concatenating label and local context features. Such a sub-network is used to learn to predict the $BoW(\delta)$ from $x^{(S)}(\tau, \delta)$. Hence, the activations of the penultimate layer, named the *coding* layer, are used to yield the CE representations. To enhance the CE, two identical sub-networks are coupled together via their coding layers for the distance learning that ensures Euclidean distance between two concepts in CE space properly reflects their semantic similarity.

[1] To distinguish from the IM learning, we apply the superscript $(S)$ to the notation of training data used in the CE learning.

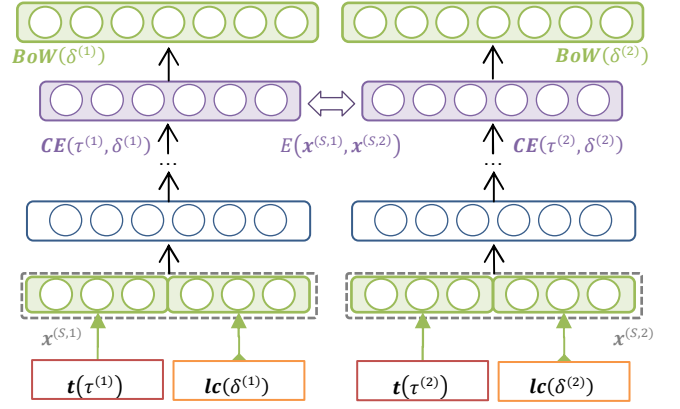

Fig. 4. *Siamese neural* architecture for concept embedding learning.

#### 3.2.3 Learning Algorithm

To learn the prediction of $y^{(S)} = BoW(\delta)$ from $x^{(S)}(\tau, \delta)$, a sub-networks is initialized with the greedy layer-wise pre-training procedure as suggested in [24]. Then, a variant of the cross-entropy loss (measuring the difference between $y^{(S)}$ and the predicted outputs, $\hat{y}^{(S)}$) is used for this learning task:

$$\mathcal{L}_P(x^{(S)}, \hat{y}^{(S)}; \Theta) = \frac{-1}{|\Gamma|}\sum_{j=1}^{|\Gamma|}(\kappa(1 + y^{(S)}[j])\log(1 + \hat{y}^{(S)}[j]) + (1-\kappa)(1 - y^{(S)}[j])\log(1 - \hat{y}^{(S)}[j])), \quad (1)$$

where $\Theta$ is a collective notation of all parameters in the sub-network, $y^{(S)}[j]$ is the $j^{th}$ element of $y^{(S)}$ and $\kappa = \frac{1}{|\Gamma|} \cdot \left|\{j: y^{(S)}[j] = 1\}_{j=1}^{|\Gamma^{(S)}|}\right|$ is a correction term that mitigates the influence of sparsity by highlighting the cost of the positive entries in $y^{(S)}$. To tackle the problem that the prediction learning is predominated by the local context features leading to improper embedding, negative examples were introduced. A negative example is synthetically generated by coupling $\delta$ randomly with a label that is not in $\delta$. Consequently, its target output is the complement of $BoW(\delta)$ by flipping the values of its entries. To avoid confusion, all examples generated from the semantic learning dataset are said as positive examples hereinafter.

The semantic distance between two concepts in the CE space, $CE(\tau^{(1)}, \delta^{(1)})$ and $CE(\tau^{(2)}, \delta^{(2)})$, is defined via Euclidian distance:

$$\mathbb{E} = \left\|CE(\tau^{(1)}, \delta^{(1)}) - CE(\tau^{(2)}, \delta^{(2)})\right\|_2. \quad (2)$$

Furthermore, the distance between the two local contexts is defined as the Kullback–Leibler (KL) divergence:

$$KL(\delta^{(1)}, \delta^{(2)}) = \sum_{c=1}^{|\Phi|}\left((lc(\delta^{(1)})[c] - lc(\delta^{(2)})[c]) \cdot \log\left(\frac{lc(\delta^{(1)})[c]}{lc(\delta^{(2)})[c]}\right)\right).$$

Based on the KL divergence, we define the similarity between two local contexts as $\mathbb{S} = e^{\frac{-\lambda}{2}KL(\delta^{(1)}, \delta^{(2)})}$. Thus, the distance learning loss is defined by

$$\mathcal{L}_S(x^{(S,1)}, x^{(S,2)}; \Theta) = I_1(\mathbb{E} - \beta(1 - \mathbb{S}))^2 \qquad (3)$$
$$+ I_2 \rho(\mathbb{E} - \beta(1 - \mathbb{S}))^2 + I_3(\mathbb{E} - \beta)^2 \mathbb{S},$$

where $\lambda$ is a positive sensitivity parameter controlling the degree to which the embedding is dominated by the context divergence, $\beta$ is a scaling parameter controlling concepts' spread over the semantics space, $I_1$, $I_2$ and $I_3$ are binary parameters specifying three possible but mutually exclusive cases regarding input to two sub-networks: both input examples are positives ($I_1$), both input examples are negative ($I_2$) and only one input example is positive ($I_3$), respectively. Finally, $\rho$ is an importance parameter that weights down the loss for $I_2 = 1$ since the accurate distance between positive and negative examples is less important than that between two positive examples.

The overall loss for the Siamese neural architecture learning is multi-objective by combining the prediction and distance learning losses in (1) and (3):

$$\mathcal{L}(x^{(S,1)}, x^{(S,2)}, y^{(S,1)}, y^{(S,2)}; \Theta) = \qquad (4)$$
$$\sum_{i=1}^{2} \mathcal{L}_P(x^{(S,i)}, y^{(S,i)}; \Theta^{(i)}) + \alpha \mathcal{L}_S(x^{(S,1)}, x^{(S,2)}; \Theta),$$

where $\alpha$ is a trade-off parameter that balances two losses and $\Theta^{(i)}$ denotes all parameters in sub-network $i$. The optimization on (4) is done with a stochastic gradient descent algorithm [25], which leads to a mini-batch based learning algorithm for this Siamese architecture [19].

After learning, one of two identical sub-networks is used as our CE model that carries out the **CE** mapping: a label $\tau$ along with its local context $\delta$ are fed to this sub-network and the coding layer outputs its CE representation, **CE**$(\tau, \delta)$. By using the CE model, any concepts in the same domain can thus be embedded in the CE space.

### 3.3 CE-Based Instance Mapping Learning

In this section, we present our approach to learning the mapping from instances to the CE representations **IM**: $\mathcal{R}^{d^{(I)}} \to S$.

#### 3.3.1 Training Example Generation

For training a model to learn *instance mapping* (IM), we need to apply the CE model described in Sect. 3.2 to an instance training dataset in order to generate the CE representations for the set of labels associated with each instance and "compress" them into target CE representation.

When there is no OOV label in $\delta = \{\tau_i\}_{i=1}^{|\delta|}$ associated with an instanace, the CE representation for $\tau_i$ in its local context $\delta$ is achieved directly via the CE model: **CE**$(\tau_i, \delta)$. In the presence of OOV labels in $\delta$, we make use of the CE nature to infer the CE representation of the OOV label from those of other in-vocabulary (IV) labels in $\delta$ [19]. As co-occurring labels in $\delta$ should be semantically coherent, the CE representation of an OOV label can be estimated as the centroid of the CE representations of co-occurring labels. Without the use of the CE model, the CE representation of an OOV label $\tau_{oov} \notin \Gamma^{(S)}$ in $\delta$ is **CE**$(\tau_{oov}, \delta) = \frac{1}{|\delta_{IV}|} \sum_{\tau_i \in \delta_{IV}} \mathbf{CE}(\tau_i, \delta_{IV})$ where $\delta_{IV}$ is the subset of $\delta$ that contains all the IV labels in $\delta$. Thus, the CE represnatations of all labels in $\Gamma^{(T)}$ associated with any training instance are achieved.

With the same considerations, we define the CE representation of a target, a "compressed" version, as

$$s = \frac{1}{|\delta|} \sum_{i=1}^{|\delta|} \mathbf{CE}(\tau_i, \delta), \qquad (5)$$

where $\delta = \{\tau_i\}_{i=1}^{|\delta|}$ is a set of labels describing instance $x$. This treatment enables us to learn the instance mapping **IM**: $\mathcal{R}^{d^{(I)}} \to \mathcal{R}^{d^{(S)}}$ with a regression model.

#### 3.3.2 SVR-Based Instance Mapping Learning

*Support vector regression* (*SVR*) [26] turns out to be a powerful tool for regression. In our work, we adopt SVR to learn a regression model. As the CE representation target **s** for an instance $x$ is multivariate, we train $d^{(S)}$ SVR models, respectively, where each SVR manages the regression from $x$ to one of $d^{(S)}$ CE features. Given an instance training dataset of $N$ examples, $\{(x_i, s_i[s])_{s=1}^{d^{(S)}}\}_{i=1}^{N}$, $\nu - SVR$ learning is defined as [27]:

Minimize $\frac{1}{2} \boldsymbol{\theta}^{(s)T} \boldsymbol{\theta}^{(s)} + C^{(s)} \left(\nu^{(s)} * \epsilon + \frac{1}{N} \sum_{i=1}^{N} \left(\xi_i^{(s)} + \xi_i^{(s)^*}\right)\right)$

subject to
$$\left(\boldsymbol{\theta}^{(s)T} \vartheta^{(s)}(x_i) + b^{(s)}\right) - s_i[s] \leq \epsilon + \xi_i^{(s)}, \qquad (6)$$
$$s_i[s] - \left(\boldsymbol{\theta}^{(s)T} \vartheta^{(s)}(x_i) + b^{(s)}\right) \leq \epsilon - \xi_i^{(s)^*},$$
$$\xi_i^{(s)}, \xi_i^{(s)^*} > 0, \ \epsilon > 0, \ i = 1, \dots, N,$$

where $\boldsymbol{\theta}^{(s)}$ and $b^{(s)}$ are linear projection parameters used to predict target values, $C^{(s)}$ is a regularization term and $0 \leq \nu^{(s)} \leq 1$ is a trade-off hyperparameter controlling $\epsilon$ in the hinge loss. $C^{(s)}$ and $\nu^{(s)}$ are chosen *a priori*. The slack variables $\xi_i^{(s)}$ and $\xi_i^{(s)^*}$ control the training error. Moreover, the function $\vartheta^{(s)}(x_i)$ is an expansion function that projects the input onto a feature space of higher dimensionality. The problem in (6) can be efficiently dealt with using the kernel trick. First, we achieve the dual formulation by using the Lagrange multiplier method [27]:

Minimize $\frac{1}{2}(\boldsymbol{\alpha}^{(s)} - \boldsymbol{\alpha}^{(s)^*})^T K^{(s)}(\boldsymbol{\alpha}^{(s)} - \boldsymbol{\alpha}^{(s)^*}) + s[s]^T(\boldsymbol{\alpha}^{(s)} - \boldsymbol{\alpha}^{(s)^*})$

subject to
$$\mathbf{1}_N^T(\boldsymbol{\alpha}^{(s)} - \boldsymbol{\alpha}^{(s)^*}) = 0$$
$$\mathbf{1}_N^T(\boldsymbol{\alpha}^{(s)} + \boldsymbol{\alpha}^{(s)^*}) \leq C^{(s)} \nu^{(s)} \qquad (7)$$
$$0 \leq \alpha_i^{(s)}, \alpha_i^{(s)^*} \leq \frac{C^{(s)}}{N}, \quad i = 1, \dots, N.$$

Here, $\alpha$ and $\alpha^*$ are Lagrange multipliers corresponding to inequality constraints in (6) and $\mathbf{1}_N$ is a $N$-dimensional vector of unit elements. $K^{(s)}_{ij} = \langle \vartheta^{(s)}(x_i), \vartheta^{(s)}(x_j) \rangle = \Bbbk^{(s)}(x_i, x_j)$ denotes a kernel, such as dot product (linear), a polynomial expansion or the *radial basis function* (RBF), and is pre-computed by using all the instance training examples. The optimization in (7) is completed via quadratic programming in it dual form [27]. We collectively denote all the optimal parameter sets for $d^{(S)}$ SVR models by $\Xi = \{\boldsymbol{\alpha}^{(s)}, \boldsymbol{\alpha}^{(s)^*}, b^{(s)}\}_{s=1}^{d^{(S)}}$. Thus, the IM regression consist-

ing of $d^{(S)}$ models is obtained by

$$IM(x;\Xi) = \left\{\sum_{i=1}^N \left(\alpha_i^{(s)*} - \alpha_i^{(s)}\right)\mathbb{k}^{(s)}(x_i,x) + b^{(s)}\right\}_{s=1}^{d^{(S)}}. \quad (8)$$

Finally, $b^{(s)}$ values are computed from (8) using one (or an average of many) training example.

### 3.4 Deployment in Multi-Label ZSL

During test, the trained IM model yields a predicted CE target $\hat{s} = IM(\hat{x};\Xi)$ for a test instance $\hat{x}$. Then, a standard semantic priming procedure [22] is applied in order to achieve the relatedness via (2) that measures the distance between $\hat{s}$ and the known embedded concepts defined by all the examples in our semantics learning and instance training datasets (c.f. Fig. 2(b)). While a label has multiple CE representations as it appears in different sets of labels used to describe different instances, the ultimate goal of Multi-label ZSL expects a single relatedness score assigned to each label. By means of the CE nature, we tackle the problem by defining the following rule: for a label $\tau_i \in \Gamma$, the relatedness between $\hat{x}$ and $\tau_i$ is measured via the minimum distance between $\hat{s}$ and any known CE representations of $\tau$, i.e., $\mathbb{E}(\hat{s},\tau_i) = min_{\delta\in\Delta}\|\hat{s} - CE(\tau_i,\delta)\|_2$. Thus, the relatedness between $\hat{x}$ and $\tau_i$ is defined by

$$\hat{l}[i|\hat{x}] = \frac{\mathbb{E}^{-1}(\hat{s},\tau_i)}{\sum_{j=1}^{|\Gamma|}\mathbb{E}^{-1}(\hat{s},\tau_j)}, \; i=1,2,\ldots,|\Gamma|. \quad (9)$$

## 4 EXPERIMENTAL SETTINGS

To evaluate our approach thoroughly, we apply it to both image and music domains. In this section, we describe datasets, experimental protocols and evaluation criteria used in this work.

### 4.1 Dataset

We use two benchmark datasets in each domain: MagTag5K [28] and Million Song Dataset (MSD) [29] for music tracks and HSUN [14] and LabelMe [30] for images.

MagTag5K is a controlled version of MagnaTune which is the result of an online annotation game where players evaluate the appropriateness of sets of labels to music tracks [31]. MagTag5K contains 5,259 music tracks annotated with a vocabulary of 136 labels. The averaging number of labels in a set of labels describing a single track, i.e., *document length*, is five in MagTag5K. MSD is a dataset of one million songs; some of which are annotated online by the crowd via *last.fm*, a crowd sharing website for users to annotate music tracks freely, where there are 218,754 MSD tracks having at least one label. MSD label usage is quite different from that of MagTag5K. This difference is illustrated in Fig. 5(a) where labels are arranged in a descending order of their MagTag5K usage.

HSUN is an image dataset of 4,367 training and 4,317 testing indoor/outdoor images. The images are annotated with a vocabulary of 107 labels and the averaging document length is 5.3 per image. LabelMe is dataset of 26,945 images annotated with 2,385 labels and the averaging document length is 7.3 per image. The difference in label usage between HSUN and LabelMe is illustrated in Fig. 5(b) with the same notation used in Fig. 5(a).

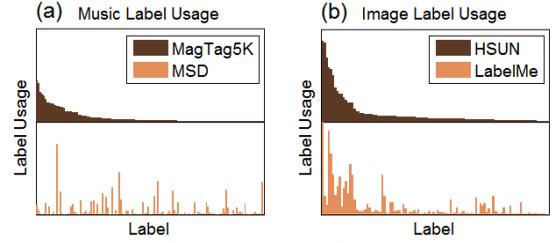

Fig. 5. Label usage distributions on different datasets. (a) Label usage in music datasets. (b) Label usage in image datasets.

It is observed that there is higher agreement between annotators on visual concepts than on musical concepts; the correlation of label usage between two image datasets is 0.75 but is only 0.07 between two music datasets. Such mismatch inevitably affects generalization of the semantics learned from one music dataset to the other.

### 4.2 Instance Input Representation

To establish the IM model, we use commonly used instance features to represent an image or a music track.

Acoustic information is extracted from a music track via short-term spectral analysis, e.g. *Echo Nest Timbre* (ENT) features [32] that characterize audio segments with 12 MFCC-like basis functions [33]. It is worth mentioning that those basis functions are kept secret by EchoNest but seamless encoding of any music track is made possible through their API [32]. Datasets such as MSD are often distributed using ENT features instead of raw music tracks in order to bypass copyright restrictions. As a result, a track $\omega$ is automatically split into $n$ segments where each segment is characterized by 12 ENT features via the API. In our experiments, the ENT features of a segment along with the 1st and 2nd derivatives constitutes the segment's feature vector $e_i$ of 36 features; and an entire track is represented with the segments features collectively, i.e., $ENT(\omega) = \{e_i\}_{i=1}^n$. ENT frames of a track are aggregated with the *Audio Bag-of-Words* (ABoW) [34], which yields a feature vector of fixed length. To achieve ABoW, a codebook $Q = [q_1,\ldots,q_{d^{(I)}}]$ of words is firstly established with Gaussian Mixture Model, where $q_j$ is a multivariate Gaussian distribution, based on a training set of instances. Each ENT frame $e_i$ is assigned its most likely code word via a 1-of-$d^{(I)}$ representational scheme:

$$\hat{f}(e_i)[j] = \begin{cases} 1 & if\; j = argmax_{j\in[1\ldots d^{(I)}]}\left(P(e_i|q_j)\right) \\ 0 & otherwise \end{cases}$$

Then, the above feature vectors for an entire track are summed to form the ABoW representation of a track:

$$ABoW(\{e_i\}_{i=1}^n) = \sum_{i=1}^n \hat{f}(e_i).$$

Finally, the feature vector is normalized to remove the effect of variable track lengths with $norm(x) = \frac{x}{\sum_{j=1}^{|x|}(x[j])}$:

$$f(\omega) = norm\left(ABoW(ENT(\omega))\right) \in \mathcal{R}^{d^{(I)}}. \quad (10)$$

In our experiment, we set the codebook size to $d^{(I)} = 128$.

Deep *Convolutional Neural Networks* (CNNs) have recently become the *de facto* image feature extractors [35]. In our experiment, we employ OverFeat [36], an off-the-shelf generic deep CNN based feature extractor trained on an image dataset with a multi-task target of object localization, detection and recognition. The CNN consists of six convolutional, two fully connected and an output layers. The output of its different hidden layers forms generic yet different image features. We use the output of the first fully connected layer to form our image representation. As a result, each image $\omega$ is initially represented by 4096 features, i.e., $\boldsymbol{OverFeat}(\omega)$. For dimension reduction, we further apply the three-layered Restricted Boltzmann Machine (RBM) [37] to $\boldsymbol{OverFeat}(\omega)$, which leads to a low dimensional representation: $\boldsymbol{RBM}(\boldsymbol{OverFeat}(\omega))$ of $d^{(I)}$ features. In our experiments, we set $d^{(I)} = 512$ based on our empirical study (see Appendix for details).

TABLE 1
INFORMATION ON DATASETS AND EXPERIMENTAL SETTINGS

| Data set | #Label | | | | #Instance | | | | | | | | |
|---|---|---|---|---|---|---|---|---|---|---|---|---|---|
| | All | Train | ZSL | OOV | All | $S_{tr}$ | $S_{val}$ | $S_{tst}$ | $IM_{tr}$ | $IM_{val}$ | $IM_{tst}$ | $OOV_{tr}$ | $OOV_{tst}$ |
| MagTag5K | 136 | 85 | 29 | 22 | 4986 | 2550±50 | 300 | 975±50 | 1914±55 | 957±55 | 953 | 2664±55 | 410 |
| MSDSub | 1305 | n/a | 29 | 1191 | 4035 | n/a | n/a | n/a | n/a | 675 | n/a | 1668 | 1692 |
| HSUN | 107 | 80 | 27 | 0 | 8684 | 2839±5 | 1527±5 | 4217 | 3074 | 2955 | 1262 | n/a | n/a |
| LabelMeSub | 651 | n/a | 27 | 544 | 4628 | n/a | n/a | n/a | n/a | 720 | n/a | 2605 | 1303 |

### 4.3 Experimental Protocol

For a thorough performance evaluation, we have designed a number of experiments in different settings and further compared our approach to COSTA [18]. To the best of our knowledge, this is the only model that uses contextualized semantics for multi-label ZSL. Other approaches are not comparable due to their technical limitations, e.g., [16], or dependence on other techniques required in their approach, e.g., semantic image segmentation has to be done prior to ZSL learning [9]. Furthermore, the work in [9] is only applicable to image domain while our experiments cover both image and music domains. In our experiments, we adopt two different settings for semantics learning.

The first setting is the same as used in COSTA [18] where a single dataset is used to simulate ZSL scenarios. As a result, the vocabulary of labels used in this dataset is randomly split into two subsets: 75% labels used for T-class labels and the remaining 25% labels used to simulate ZSL-class labels. We name this setting *within-corpus test* (WCT).

In WCT, we use multi-trial *cross-validation* (CV) for performance evaluation. In each CV trial, a dataset is randomly split into two data subsets: $D_1$ and $D_2$. All the annotation documents of instances in $D_1$ are used for semantic learning. As a result, $D_1$ is further divided into two subsets $S_{tr}$ and $S_{val}$ that are used for parameter estimation as well as searching for optimal hyperparameters and avoiding over-fitting, respectively. For the IM learning, all the instances of T-class labels in $D_1$ and $D_2$ constitute the training and validation sets, $IM_{tr}$ and $IM_{val}$, respectively. Consequently, all instances with at least one ZSL-class label in the dataset (i.e., $D_1$ and $D_2$) form the test set, $IM_{tst}$. In our experiments, we conduct the WCT experiments on MagTag5K and HSUN. For MagTag5K, we follow the dataset splitting suggested in [28]: the number of instances in $D_1$ is twice of that in $D_2$, and $D_1$ is randomly split into $S_{tr}$ and $S_{val}$ as listed in Table 1. In HSUN, all the instances were pre-split into training and test sets [14]. Thus, we follow this setting by using the training data for learning semantic representations and regressors and conducting testing on the test data. Table 1 contains the information on datasets and their split subsets described above, where three trials of CV are conducted. For proof of concept, we further employ MagTag5K to simulate an OOV scenario by reserving 22 labels as OOV labels; all the annotation documents containing any of 22 OOV labels are not used in the CE learning. For the IM learning, however, we used all the instances in $IM_{tr}$ plus those instances described using only T-class and OOV labels to form the training set, $OOV_{tr}$. Accordingly all the remaining instances associated with ZSL-class and OOV labels constitute the corresponding OOV test set, $OOV_{tst}$, as listed in Table 1.

Unlike previous works, we further create an alternative setting: for two datasets in the same domain, the semantics learning model is trained on one dataset and then the learned semantics is directly applied to the other for multi-label ZSL. We refer this setting as to *cross-corpora test* (CCT). Thus, CCT provides an effective way to evaluate the generalization of learned semantics. In our CCT experiments, we use MagTag5K and HSUN for semantics learning, and the CE models achieved are applied to instance mapping learning on MSD and LabelMe, respectively. As there are much more labels used in MSD and LabelMe than those in MagTag5K and HSUN, we have to use subsets of MSD and LabelMe, MSDSub and LabelMeSub, where each instance is associated with in-vocabulary labels of MagTag5K and HSUN and/or up to two OOV labels. This setting is due to the fact that concepts defined by OOV labels have to be approximated with their co-occurring in-vocabulary labels and a predominate number of OOV labels in an annotation document inevitably lead to inaccurate approximation.

In the CCT, T-class and ZSL-class labels specified in our WCT remain, and the IM learning follows the same convention: only instances of T-class and OOV labels are allowed to be used in training and those containing ZSL-class labels are retained for test. It is worth stating that there are a very limited number of instances of only in-vocabulary labels (i.e., those used in MagTag5K and HSUN) but a vast majority of instances with OOV labels in MSDSub and LabelMeSub. In the CCT, we do not distinguish between these two types of instances. Once again, the same CV procedure used in the WCT is applied to the IM learning. Thus, a dataset is split into training, validation and test subset, $OOV_{tr}$, $IM_{val}$ and $OOV_{tst}$, as shown in Table 1.

To see performance in different scenarios clearly, we report the performance of a multi-label ZSL model separately based on various test instance subsets where instances are associated with different types of labels:

**Training Labels.** Test instances are associated with only in-vocabulary T-class labels in $\Gamma^{(T)} \cap \Gamma^{(S)}$. This corresponds to the traditional multi-label classification [8] but is not the main focus in this work.

**ZSL Labels.** Test instances are associated with at least one ZSL-class label in $\Gamma^{(S)}$. In this circumstance, a model has to deal with test data of ZSL-class labels, a typical ZSL evaluation scenario.

**All Labels.** Test instances are associated with all kinds of labels including T-class, ZSL-class and OOV labels. In reality, a model has to deal with this real world scenario.

**OOV Labels.** This evaluation focuses on the performance of the OOV labels only. Note that this evaluation is only applicable to our model as the existing multi-label ZSL models including COSTA [18] have yet to take this into account.

## 5 EVALUATION

In this section, we first describe our evaluation criteria and report the results on different experimental settings.

### 5.1 Evaluation Criteria

In general, multi-label classification can be evaluated in two paradigms: *example-based* and *concept-based* [38]. The example-based evaluation assesses the ability of a model in predicting a set of suitable labels for a test instance, while the concept-based evaluation examines the capability of a model in correctly identifying the applicability of individual labels to test instances. Unlike COSTA [18] which used only the concept-based evaluation, we adopt both evaluation criteria in our experiments.

Given a test instance $\hat{x}$, a model yields the ranked relatedness scores to all known labels: $\hat{l} = \{\hat{l}[i|\hat{x}]\}_{i=1}^{|\Gamma|}$ where $\hat{l}[i|\hat{x}] \geq \hat{l}[j|\hat{x}]$ if $i < j$, as described in Sect. 3.4. In the example-based evaluation, we first measure the precision at $k$ [39, pp. 151–162], i.e., the proportion of correctly predicted labels in the top $k$ positions in $\hat{l}$: $P@k(k; \delta, \hat{l}) = \frac{1}{k} \cdot |\delta \cap \hat{l}[1, ..., k]|$ where $\delta$ is the ground-truth label set of $\hat{x}$. To remove the effect of variable ground-truth document length, all $P@k$ values are further normalized based on the actual document length, which leads to the *Mean Average Precision* (MAP):

$$MAP(\delta, \hat{l}) = \frac{1}{|\delta|} \cdot \sum_{i=1}^{|\delta|} P@k(i; \delta, \hat{l}). \quad (11)$$

Hereinafter, we refer to this evaluation measure as *example-based MAP* (E-MAP).

The concept-based evaluation is performed by evaluating the prediction of a specific label in all associated instances. Given one label $\tau \in \Gamma$ which is predicted by a model to associate with a number of test instances, collectively denoted by $\gamma$, we can achieve a ranked list $\hat{l}^*$ where test instances in $\gamma$ are arranged in the descending order in terms of their relatedness scores, i.e., $\hat{l}^*[i] \geq \hat{l}^*[j]$ if $i < j$. The resultant list is then evaluated against the ground-truth via the Precision-Recall curves [38], where the precision at $k$ is the same as defined for E-MAP and the recall at level $k$ is the proportion of correctly predicted instances in the top $k$ positions in $\hat{l}^*$ in terms of the total number of instances in $\gamma$, i.e., $R@k(k; \gamma, \hat{l}^*) = \frac{1}{|\gamma|} \cdot |\gamma \cap \hat{l}^*[1, ..., k]|$. The resulting Precision-Recall curve is aggregated by averaging the precision values at the 11 standard recall levels $r \in \{0.0, 0.1, ..., 1.0\}$:

$$AP(\gamma, \hat{l}) = \frac{1}{11} \sum_{j \in r} \{P@k(i; \gamma, \hat{l})\}_{R@k(i; \gamma, \hat{l}) = j}. \quad (12)$$

Hereinafter, we refer to $AP(\gamma, \hat{l})$ as the *concept-based MAP* (C-MAP).

In our CE-ML-ZSL, the output relatedness scores can be treated as posterior probability: $P(\tau_i|\hat{x}) = \hat{l}[i|\hat{x}]$. However, the raw scores achieved by COSTA [18] are achieved for each label independently, which can be viewed as a pseudo-likelihood of an example given a label, i.e., $L(\hat{x}|\tau_i)$. To make both approaches comparable, we apply normalization and to convert COSTA score to $P(\tau_i|\hat{x}) = \frac{L(\hat{x}|\tau_i)}{(\sum_j L(\hat{x}|\tau_j))} \times \frac{P(\tau_i)}{P(\hat{x})}$ where $P(\tau_i)$ is estimated based on a semantic learning data subset $S_{tr}$ and $P(\hat{x})$ are assumed to be the same for all $\hat{x}$. In addition, we employ the RBF kernel instead of the suggested linear kernel COSTA [18] in our experiments since our empirical studies suggest that the non-linear kernel leads to better performance.

### 5.2 Results on Learning

#### 5.2.1 Results on CE Learning

During CE learning, we set the number of topics used in context modeling with the hierarchical Dirichlet process [40], which yields 19 and 30 topics for MagTag5K and HSUN, respectively. The optimal hyperparameters in the deep sub-networks are found via grid search based on the CV described in Sect. 4.3. As a result, the optimal sub-network in the Siamese architecture has a structure: $input \to 100 \to 100 \to d^{(S)} \to output$ for MagTag5K and $input \to 200 \to 200 \to d^{(S)} \to output$ for HSUN. We set $\rho = 0.5$ and $\beta = \sqrt{d^{(S)}}$ in (3), $\alpha = 1$ in (4) for both datasets. Initial learning rates are set to $10^{-4}$ for MagTag5K and $5 \times 10^{-6}$ for HSUN and the learning rates are decayed with a factor of 0.95 each 200 epochs.

In this experiment, we would evaluate the performance of our CE model by assuming that the regression done by an IM model is error-free. In other words, we use the ground-truth target of a test instance in $IM_{tst}$ achieved via (5) to evaluate the CE learning with E-MAP and C-MAP to see if the CE representations are effective for CE-ML-ZSL. Also this is the maximum limit that our CE-ML-ZSL can yield in performance and hence can be used as a reference against the test results in real scenarios.

## Fig. 6 WCT E-MAP and C-MAP performance on MagTag5K

| | E–MAP | | C–MAP |
|---|---|---|---|
| | | Training Labels | |
| | 67.40±0.09 | $d^{(S)}=10$ | 41.08±0.11 |
| | 86.31±2.68 | $d^{(S)}=50$ | 70.46±3.97 |
| | 91.96±1.66 | $d^{(S)}=200$ | 78.01±1.53 |
| | | ZSL Labels | |
| | 67.03±0.10 | $d^{(S)}=10$ | 40.40±0.86 |
| | 86.68±2.51 | $d^{(S)}=50$ | 71.63±3.71 |
| | 92.47±1.43 | $d^{(S)}=200$ | 79.27±1.91 |

Fig. 6. WCT E-MAP and C-MAP performance (mean and standard error) on MagTag5K on condition that the IM model is error-free. The notation in this figure is used in all the remaining figures.

## Fig. 7 WCT results on HSUN

| | E–MAP | | C–MAP |
|---|---|---|---|
| | | Training Labels | |
| | 36.28±0.77 | $d^{(S)}=10$ | 26.13±0.40 |
| | 62.67±0.54 | $d^{(S)}=50$ | 42.38±0.69 |
| | 70.53±0.87 | $d^{(S)}=200$ | 46.38±0.47 |
| | | ZSL Labels | |
| | 37.52±0.64 | $d^{(S)}=10$ | 26.22±1.38 |
| | 62.43±0.32 | $d^{(S)}=50$ | 43.93±0.39 |
| | 69.65±0.77 | $d^{(S)}=200$ | 47.94±0.16 |

Fig. 7. WCT results on HSUN on condition that the IM model is error-free.

Figs 6 and 7 show the performance corresponding to different dimensions of the CE space as well as two different types of labels on MagTag5K and HSUN, respectively. It is observed from Figs 6 and 7 that the dimensionality of the CE space, $d^{(S)}$, significantly affects the performance on two datasets but the CE model generalizes the learning semantics well given the fact that the performance on two different types of labels is quite similar. In general, a higher CE dimension leads to better performance probably due to the fact that a higher dimensional CE space has larger room to allow concepts to be embedded properly as required by CE learning. The results shown in Figs 6 and 7 strongly suggest that the CE representation is effective in modeling the complex semantics required by multi-label ZSL.

### 5.2.2 Results on IM Learning

For the IM learning, we use RBF kernel $\nu - SVR$ to build up a regressor to map instance input feature vectors to their CE targets. By using the CV, the optimal hyperparameters of $\nu$, $C$ and $\gamma_{RBF}$ in (7) is again found via grid search in LIBSVM [41]. In our experiments, we observe that the optimal hyperparameters depend on the dimensionality of the CE space, and are retained within a range, $\nu \in [0.1, 0.4]$, $C = 1$ and $\gamma_{RBF} = 1$ for all $d^{(S)}$ dimensions.

The IM model is evaluated by measuring the averaging error, $\hat{\varepsilon}$, incurred by regression on a test dataset, $IM_{tst}$: $\hat{\varepsilon} = \frac{1}{|IM_{tst}|} \cdot \sum_{\hat{x} \in IM_{tst}} \| \hat{s}(\hat{x}, \delta) - s(\hat{x}, \delta) \|_2$ where $\delta$ is the ground-truth label set of a test instance $\hat{x}$, $\hat{s}(\hat{x}, \delta) = IM(\hat{x}; \Xi)$ and $s(\hat{x}, \delta) = \frac{1}{|\delta|} \sum_{\tau \in \delta} CE(\tau, \delta)$. Moreover, we introduce the scattering to form another regression measurement. The scattering is defined by averaging all CE distances between known concepts to reflect information on the distribution of known concepts in the CE space. Using this statistical property, we further define the relative regression error by $\varepsilon = \hat{\varepsilon} / s$, where

## TABLE 2
REGRESSION PERFORMANCE OF IM MODEL.

| Data set | $d^{(S)}$ | Measurement | | |
|---|---|---|---|---|
| | | $s$ | $\hat{\varepsilon}$ | $\varepsilon$ |
| MagTag5K | 10 | 2.58 | 1.00 | 0.39 |
| | 50 | 25.27 | 3.35 | 0.13 |
| | 200 | 173.15 | 8.13 | **0.05** |
| HSUN | 10 | 8.56 | 1.35 | 0.16 |
| | 50 | 28.42 | 3.22 | 0.11 |
| | 200 | 171.93 | 7.06 | **0.04** |

$s = \frac{1}{\left(\sum_{\tau \in \Gamma(S)} N_\tau^{(S)}\right)^2} \sum_{\delta, \delta' \in \Delta} \sum_{\tau \in \delta, \tau' \in \delta'} \| CE(\tau, \delta) - CE(\tau', \delta') \|_2$ is achieved based on all the known concepts defined in the semantic learning data set (c.f. Sect. 3.1). Intuitively, the smaller the value of $\varepsilon$, the better the IM model performs since it implies that ground-truth labels of test instances are more likely to be found via semantic priming.

Table 2 lists the regression performance of the IM models corresponding to different dimensions of the CE space. From Table 2, it is evident that the best performance corresponds to the CE space of a dimension, $d^{(S)} = 200$. We hence use this 200-dimensional CE representation in all the experiments described in the sequel.

### 5.3 WCT Results

Now we report the experiment results in WCT, as described in Sect. 4.3, and compare our CE-ML-ZSL model to COSTA with their original setting [18]. In COSTA, the test on Training Labels is boiled down to the traditional multi-label classification. For the test on ZSL Labels, it first predicts T-class labels and then feeds the T-class prediction to linear regressors to predict ZSL-class labels.

Figs 8 and 9 illustrate the test results on MagTag5K and HSUN, respectively, in terms of two types of labels. It is evident that the performance of COSTA is degraded in predicting ZSL-class labels, as shown in results on ZSL Labels in Figs 8 and 9. It is worth mentioning that COSTA was evaluated with C-MAP in [18] and the results shown here are consistent with those in [18]. In contrast, our CE-ML-ZSL outperforms COSTA in all different types of labels on two datasets with statistical significance (Student's t-test p-value<0.05) except in one case: C-MAP of Training Labels on HSUN where the two models achieve comparable results (no statistical advantage to either model). In particular, our model achieves similar performance in predicting T-class and ZSL-class labels. In addition, it is observed from Fig. 8 that there is a much higher standard error generated by COSTA than ours on MagTag5K in E-MAP. To a great extent, this caused by the limitation of COSTA that predicts all the T-class labels independently without considering the coherence in a specific set of labels associated with an instance sufficiently. Thanks to our CE model that takes contextualized semantics into account, our model is insensitive to the CV setting and performs stably as is reflected in its E-MAP performance shown in Fig. 8.

| | E–MAP | | | C–MAP |
|---|---|---|---|---|
| | 21.66±7.24 | Training Labels COSTA | 13.16±0.25 | |
| | 47.20±1.69 | CE–ML–ZSL | 18.86±0.34 | |
| | | ZSL Labels | | |
| | 18.96±9.18 | COSTA | 11.64±0.06 | |
| | 45.64±1.51 | CE–ML–ZSL | 18.22±0.28 | |

Fig. 8. WCT results on the IM test set of MagTag5K.

| | E–MAP | | | C–MAP |
|---|---|---|---|---|
| | 23.18±0.05 | Training Labels COSTA | 24.45±0.13 | |
| | 58.20±0.29 | CE–ML–ZSL | 23.11±0.54 | |
| | | ZSL Labels | | |
| | 25.90±0.05 | COSTA | 20.40±0.10 | |
| | 56.06±0.16 | CE–ML–ZSL | 22.70±0.33 | |

Fig. 9. WCT results on the IM test set of HSUN.

| | E–MAP | | | C–MAP |
|---|---|---|---|---|
| | 29.16±9.73 | Training Labels COSTA | 12.90±0.20 | |
| | 48.50±1.50 | CE–ML–ZSL | 17.98±0.22 | |
| | | ZSL Labels | | |
| | 24.92±12.32 | COSTA | 11.33±0.32 | |
| | 47.40±1.39 | CE–ML–ZSL | 16.66±0.23 | |
| | | All Labels | | |
| | 23.58±0.17 | CE–ML–ZSL | 16.42±0.24 | |
| | | OOV Labels | | |
| | 21.51±1.76 | CE–ML–ZSL | 15.20±0.35 | |

Fig. 10. WCT results on the OOV test set of MagTag5K.

| | E–MAP | | | C–MAP |
|---|---|---|---|---|
| | 52.62±2.15 | Training Labels | 19.90±0.97 | |
| | 64.50±3.41 | ZSL Labels | 23.90±0.60 | |
| | 44.82±4.66 | All Labels | 2.39±0.04 | |
| | 0.14±0.02 | OOV Labels | 0.53±0.01 | |

Fig. 11. CCT results on MSDSub on condition that the IM model is error-free.

| | E–MAP | | | C–MAP |
|---|---|---|---|---|
| | 37.03±6.23 | Training Labels | 35.58±0.72 | |
| | 52.38±9.31 | ZSL Labels | 38.55±0.80 | |
| | 40.43±7.30 | All Labels | 6.96±0.23 | |
| | 2.66±0.32 | OOV Labels | 1.35±0.08 | |

Fig. 12. CCT results on LabelMeSub on condition that the IM model is error-free.

In presence of OOV labels, COSTA simply ignores such labels in their treatment [18]. In other words, COSTA only predicts in-vocabulary ZSL-class labels based on T-class labels. Hence, we follow their experimental protocol in OOV test on MagTag5K. Fig. 10 illustrates the results on the OOV test set of MagTag5K. It is observed that COSTA achieves slightly higher mean E-MAP values along with larger standard errors on this test dataset than its own performance on the IM test dataset shown in Fig. 8 as OOV labels do not affect the prediction of in-vocabulary labels in COSTA. Similarly, our model also slightly improves its E-MAP performance in predicting in-vocabulary T-class and ZSL-class labels on this test dataset as shown in Fig. 10 where it is seen that larger standard errors made by COSTA results in a reduction in the statistical significance on the difference between the two models in E-MAP (Student's t-test p-value<0.15). The existence of OOV labels in the ground-truth label set used to describe an instance slightly decreases the C-MAP performance of both models on Training and ZSL Labels but our model still outperforms COSTA. In C-MAP, the relevant OOV labels have to be considered but the concepts framed by such labels are either ignored in COSTA or approximated in our model. A lack of the accurate semantic information on OOV labels is responsible for the degraded performance (c.f. Figs 8 and 10). Nevertheless, our model still results in statistically significant (Student's t-test p-value<0.05) improvements over COSTA. As shown in Fig. 10, our model yields the performance on All Labels similar to that of ZSL Labels, which demonstrate the effectiveness of our model in presence of OOV labels. In particular, it is evident from Fig. 10 that our model correctly predicts a number of ground-truth OOV labels associated with instances.

Here, we emphasize that other multi-label ZSL models including COSTA cannot predict any OOV labels associated with a test instance while our model works well as shown in Fig. 10.

### 5.4 CCT Results

By using the same rubric used in Sect. 5.2 and 5.3, we report experimental results on CCT where the CE model trained on a dataset is used in another, different dataset for IM learning as described in Sect. 4.3.

We first evaluate the generalization of CE models trained on MagTag5K and HSUN. Assume that the IM model is error free. Fig. 11 shows the performance on MSDSub based on the CE model trained on MagTag5K, while Fig. 12 illustrates the performance on LabelMeSub based on the CE model trained on HSUN. It is observed from Figs 11 and 12 that the learned semantics is transferable to a great extent although the E-MAP and C-MAP performance drops considerably in comparison to that on their source datasets under WCT as shown in Figs 6 and 7. In particular, the E-MAP results vary between different CV trials as suggested by large standard errors. As seen in Fig. 6, the label usage is quite different across different datasets even in the same domain. The disparity of label usages accounts for the degraded results, which is clearly evident especially for two music datasets as shown in Fig. 11. As one of distinguishing CCT characteristics, there are many OOV labels not appearing in CE learning. We further evaluate the performance on All Labels and OOV Labels and the results are shown in Figs 11 and 12. It is seen that E-MAP is high for All Labels but C-MAP is low. In fact, the E-MAP considers the predictions of suitable groups of labels which might include few OOV labels, while C-MAP is averaged over all labels. Thus, C-MAP for an OOV label is naturally low due to a lack of information surrounding the intended concept defined by an OOV label. It is also observed that the performance on OOV Labels is extremely low. This experiment exhibits the great challenge in predicting one or two OOV labels correctly from a large OOV vocabulary, e.g., there are 1,191 and 544 OOV labels in music and image domains, respectively. To the best of our knowledge, our work here is the very first attempt, which will be discussed later on.

| E-MAP | Training Labels | C-MAP |
|---|---|---|
| 0.05±0.05 | COSTA | 2.30±0.10 |
| 8.34±0.98 | CE-ML-ZSL | 3.71±0.02 |
| | ZSL Labels | |
| 0.17±0.17 | COSTA | 3.27±0.08 |
| 7.67±0.76 | CE-ML-ZSL | 5.22±0.07 |
| | All Labels | |
| 0.74±0.05 | CE-ML-ZSL | 0.71±0.01 |
| | OOV Labels | |
| 0.30±0.03 | CE-ML-ZSL | 0.34±0.01 |

Fig. 13. CCT results on the IM test set of MSDSub.

| E-MAP | Training Labels | C-MAP |
|---|---|---|
| 4.03±0.05 | COSTA | 7.18±0.17 |
| 14.72±1.27 | CE-ML-ZSL | 12.04±0.61 |
| | ZSL Labels | |
| 4.24±0.11 | COSTA | 7.32±0.17 |
| 14.16±1.11 | CE-ML-ZSL | 13.54±0.51 |
| | All Labels | |
| 2.72±0.10 | CE-ML-ZSL | 2.99±0.07 |
| | OOV Labels | |
| 1.84±0.18 | CE-ML-ZSL | 1.15±0.03 |

Fig. 14. CCT results on the IM test set of LabelMeSub.

Now, we report the performance of the IM models in the CCT setting, as described in Sect. 4.3, on the IM test datasets. For COSTA, in this setting, the classifiers for predicting T-class labels are trained on the target datasets, i.e., MSDSub and LabelMeSub, but the regressors and prior likelihood estimations remain the same as established based on the source datasets, i.e., MagTag5K and HSUN, respectively. Figs 13 and 14 show the E-MAP and C-MAP performance on test instances in terms of different types of labels. It is observed that both models yield better performance on the image domain than that on the music domain due to the fact there is a higher agreement between people on visual concepts than musical ones. While the performance in CCT is generally less satisfactory, our CE-ML-ZSL always outperforms COSTA considerably in all different test instance subsets and with the statistical significance (Student's t-test p-value<0.05). Results shown in Figs 13 and 14 suggest that the regression task involved in IM learning is extremely challenging but, to some extent, semantics learned from one dataset may be generalized to another in particular when there is little disparity between two datasets in the same domain. For our CE-ML-ZSL, we further evaluate its performance on OOV Labels and show the results in Figs 13 and 14. As expected, the performance on All Labels and on OOV Labels is quite low but still exhibits our first attempt by using a novel way to deal with OOV labels in a real word scenario.

In summary, the experimental results reported in this section clearly demonstrate that our CE-ML-ZSL outperforms COSTA, a state-of-the-art multi-label ZSL approach, in different settings.

# 6 DISCUSSION

In this section, we discuss issues arising from our CE-ML-ZSL framework and relate ours to previous works.

In CE-ML-ZSL, we deal with multi-label ZSL by means of our very recent work in learning contextualized semantics from co-occurring labels [19]. Modeling such contextualized semantics underlying a set of co-occurring labels leads to concept embedding (CE), where a label has multiple representations when co-occurring with different labels. This semantic representation models the complex semantics formulated jointly by a set of co-occurring labels and its learning is independent of instance input features. Hence, it complies with the generic multi-label ZSL requirement. Learned semantics from a label dataset may be applied to different datasets for multi-label ZSL provided there are similar label usages across concerned datasets. As our CE-ML-ZSL deals with T-class and ZSL-class labels in a similar manner, our approach may lead to the better performance on test instances in the presence of ZSL-class labels than that on those associated with only T-class labels as demonstrated in our experiment results reported in Sect. 5. Furthermore, the nature of the CE representation enables us to learn mapping from instance input features to semantic space efficiently, i.e., in a similar manner used in single-label ZSL, and to deal with OOV labels without re-training the CE model. Hence, our approach does not suffer from the exhaustive search problem in deployment, which is encountered in [16].

Nevertheless, the current implementation of CE-ML-ZSL is subject to limitations. On the one hand, CE learning is carried out by a Siamese deep architecture of which training suffers from a high computational burden, a common weakness of deep neural networks [42], especially for a large vocabulary of many labels, e.g., MSD. Due to the limitation of our computing facility, the CE learning in our experiments has to be confined to datasets of a moderate size as well as the dimensionality of CE space is limited to $d^{(S)} = 200$. This results in a serious difficulty in conducting CCT experiments as it does not seem sensible to expect that semantics learned from a small dataset to generalize well to a large dataset of many more labels and resultant concepts. Therefore, scalable CE learning is an issue that has to be addressed in CE-ML-ZSL. On the other hand, like most of ZSL approaches [2]–[7], a multivariate regressor has to be learned in CE-ML-ZSL to map instance input features to its corresponding semantic representation. Due to the well-known semantic gap [43], the regression learning appears extremely difficult in our work including those not reported due to the limited space here, which has also been reported in previous studies [44]. In reality, noisy labeling and label missing in training data exacerbate this problem. In general, a satisfactory solution relies on both proper instance input features and powerful regressors. In our experiments, we simply adopt those commonly used input features for a proof of concept, while the use of $\nu - SVR$ is a trade-off between the performance and computational efficiency. Hence, it is yet another issue for CE-ML-ZSL to explore proper input features in different domains and powerful regressors.

Unlike the previous works in multi-label ZSL, we make the very first attempt in investigating real scenarios,

i.e., in presence of OOV labels and the requirement of transferring semantics from a label dataset to another, via novel experimental settings. Our results on MagTag5K suggest that it is promising to predict OOV labels with the CE representation without involving re-training learned semantics, and the CCT experiments reveal that it is essential to have similar label usages for success in transferring learned semantics. We anticipate that such experimental settings may trigger the interest in research into multi-label ZSL by considering various real world scenarios.

The use of CE representation becomes the most salient feature in distinguishing our CE-ML-ZSL framework from existing multi-label ZSL approaches. While our approach uses only contextualized semantics learned from co-occurring labels, most of the works [9], [16] have to rely on linguistic semantics learned by analyzing syntactic contexts of words in a natural language. However, a set of labels used to describe an instance are neither arranged in a syntactic order nor limited to words in natural languages. In addition, our CE representation is significantly different from the semantics used in COSTA [18] since they take neither correlation between co-occurring labels nor the polysemantic aspect of a label into account. Apart from the above issue in modeling complex semantics accurately, the semantics used in existing multi-label ZSL approaches inevitably result in difficulties in dealing with OOV labels and mapping instance input features to semantic space efficiently.

While our CE-ML-ZSL is a generic framework that allows being applied to multi-label ZSL tasks in different domains, it does not carry out useful functions fulfilled by some domain-specific approaches. For example, the multi-label ZSL model proposed in [9] is specific to image domain for zero-shot multi-label object classification. Based on an image segmentation mechanism, it not only predicts a set of labels associated with an image but also localizes objects with their corresponding labels. Hence, our approach is subject to limitations in some specific applications.

As same as done by most of existing ZSL approaches, e.g., [2]–[7], [18], our CE-ML-ZSL directly maps instance input features to semantic space to carry out ZSL. Due to the semantic gap, the regression task involved in the mapping learning is extremely difficult. To bridge the semantic gap, an alternative idea is learning an intermediate-level latent embedding simultaneously from input features of instances and their labels, e.g., [4], [45]. Although such approaches are only suitable for specific applications and does not address multi-label ZSL, we believe it could provide an alternative insight into multi-label ZSL. Thus, it might result in a different research direction that could improve our CE-ML-ZSL framework.

## 7 CONCLUSION

In this paper, we have presented a novel yet generic multi-label ZSL framework, CE-ML-ZSL. In CE-ML-ZSL, we emphasize the importance of modeling complex semantics underlying a set of co-occurring labels used to describe an instance and come up with a solution by means of the CE learning [19]. The resultant CE representation forms polysemantic representations of a label in context of its co-occurring labels that are together used to describe an instance. Thanks to the CE representation, we can establish a mapping from the input features of an instance onto the semantic space very efficiently in a similar manner used in single-label ZSL. The experimental results via the thorough evaluation with different settings, evaluation criteria and the comparison to COSTA [18] suggest that our CE-ML-ZSL yields favorable multi-label ZSL performance in music and image domains. Furthermore, we have made first attempts in dealing with OOV labels without re-training the semantics learning model and conducting the cross-corpora test. Such attempts could inspire deepening multi-label ZSL researches so that developed techniques can effectively tackle different multi-label ZSL scenarios in the real world.

In our ongoing work, we are going to address issues arising from this research including scalability of CE learning, effective approaches to mapping instance content features onto semantic space and exploring real applications of our CE-ML-ZSL in multimedia information retrieval towards bridging the semantics gap between machine extractable features and human level understanding.

## APPENDIX: IMAGE FEATURE SET SELECTION

In this appendix, we describe how we achieve the image feature set used in our experiments presented in the main text. The purpose of this image feature set selection experiment is two-fold: a) making a fair comparison to COSTA [18] by finding out an alternative feature set that leads to the performance comparable to that reported in [18] given the fact that the handcrafted features were used in [18] but neither the details of feature extraction nor their code/data are available publicly; and b) exploiting the state-of-the-art feature learning techniques to find out proper features in order to facilitate our instance mapping learning.

### A.1 OverFeat-Based Image Representations

OverFeat [36] is a generic deep CNN trained on images for object classification, localization or detection. The network consists of six convolutional layers followed by two fully-connected layers of the same size and ended to the output layer. Thus, the outputs of its hidden layers lead to various high-level image representations. The output of the last convolutional layer, dubbed F19, forms a localized image descriptor, a vector of 26,000 features. The first fully-connected layer functions to detect global patterns existing in F19. This layer consists of 4,096 neurons and their outputs, dubbed F21, constitute a feature vector. Based on F21, the second fully-connected layer further facilitates the output layer to accomplish the ultimate goal. Hence, the output of this penultimate layer, dubbed F23, lead to a task-oriented global descriptor, a vector 4,096 features. Consequently, F19, F21 and F23 re-sult in three different image feature sets and each could be used for multi-label classification concerned in this study.

One of the latest works [37] suggests that applying a nonlinear dimension reduction technique to image representations resulting from OverFeat, or similar image descriptors, may yield the improved performance in image retrieval. In [37], a deep RBM is employed to generate the unsupervised triple hashing (UTH) of images where the output of the last hidden layer of fewer neurons than the dimension of input space forms a low-dimensional representation. To obtain the UTH, a deep RBM is pre-trained in a layer-wise greedy fashion and then fine-tuned to preserve the distances of input space in the low-dimensional latent space.

This representation was found to perform well for an image retrieval task. We preserve the parameterization reported to perform best in [37].

### A.2 Experimental Setting

As we have three candidate image representations and each can be reduced to different low dimensions via pre-trained RBM and the UTH after fine-tuning the RBM, we have conducted experiments to select the optimal low-dimensional feature set via finding the feature set that yields a lowest dimension representation with the minimum accuracy loss in our multi-label classification task.

Following the setting used in COSTA [18], we train a set of independent binary classifiers to predict Training Labels by using three different image feature sets, F19, F21 and F23, on the IM training set $IM_{tr}$ of HSUN (c.f. Sect. 4.3 and Table 1 in the main text), respectively. Then, each candidate representation is evaluated on the IM validate set $IM_{val}$ with the three-fold CV. This setting can be viewed as a simplified version of our WCT evaluation setting described in Sect. 4.3 by taking only the validation accuracy into account on different image feature sets and subsequently a number of low-dimensional representations of the optimal feature set. In our experiment, we retain the parameterization that yields the best performance in [37] to obtain a low-dimension representation.

Following the setting used in COSTA [18], we train a set of independent binary classifiers to predict Training Labels by using three different image feature sets, F19, F21 and F23, on the IM training set $IM_{tr}$ of HSUN (c.f. Sect. 4.3 and Table 1 in the main text), respectively. Then, each candidate representation is evaluated on the IM validate set $IM_{val}$ with the three-fold CV. This setting can be viewed as a simplified version of our WCT evaluation setting described in Sect. 4.3 by taking only the validation accuracy into account on different image feature sets and subsequently a number of low-dimensional representations of the optimal feature set. In our experiment, we retain the parameterization that yields the best performance in [37] to obtain a low-dimension representation.

### A.3 Results on Image Feature Set Selection

Following the same notation used in the main text, we report the performance in Fig A.1 and Fig A.2 by using one evaluation criterion: C-MAP (c.f. Sect 5.1 of the main text).

| OverFeat | C-MAP |
|---|---|
| F19(26000) | 27.19±0.12 |
| F21(4096) | 29.91±0.15 |
| F23(4096) | 28.64±0.04 |

Fig. A.1. Accuracies resulting from F19, F21 and F23 image feature sets on IM Validation set of HSUN. The number of features in a representation is listed in parenthesis.

| RBM | C-MAP |
|---|---|
| RBM(256) | 26.44±0.05 |
| RBM(512) | 26.96±0.05 |
| **UTH** | |
| UTH(256) | 25.23±0.09 |
| UTH(512) | 25.17±0.02 |

Fig. A.2. Accuracies resulting different low dimensional representations of F21 yielded by the pre-trained RBM and the UTH on the IM Validation set of HSUN.

It is observed from Fig. A.1 that F21 leads to almost identical accuracy to the best result, 28.3%, reported in [18]. F23 comes seconds with satisfactory results. For comparative studies, we are most interested in comparing our model to COSTA model [18]. For a fair comparison, we choose F21 to be the "optimal" feature set based on the performance in C-MAP only.

By setting different numbers of hidden neurons in the representation layer in a deep RBM, we train the deep RBM on F21 image feature sets extracted from training data of HSUN, which leads to two low-dimensional representations, e.g., the pre-trained RBM and the UTH, for each setting. It is evident from Fig. A.2 that accuracies on all low-dimensional representations have been decreased slightly due to dimensionality reduction. Furthermore, the fine-tuned RBM, i.e., UTH, underperforms the pre-trained RBM. By comparing different dimensions, we find RBM (512) outperforms RBM (256).

Based on experimental results reported above, we choose the low dimensional representation of F21 image feature set yielded by the pre-trained RBM, i.e., RBM (512), to be the representation of images in HSUN used in our WCT and CCT experiments.